\documentclass[letterpaper, 10 pt, conference]{ieeeconf}  

\IEEEoverridecommandlockouts                              

\usepackage[noadjust]{cite}
\usepackage{color}
\usepackage{subcaption}
\usepackage{graphics} 
\usepackage{amsmath} 
\usepackage{amssymb}  
\usepackage{booktabs}
\usepackage{graphicx}
\usepackage{url}
\usepackage[capitalise]{cleveref}

\title{\LARGE \bf
Striking the Right Balance: Recall Loss for Semantic Segmentation
}

\author{ Junjiao Tian*, Niluthpol Mithun$\dagger$, Zachary Seymour$\dagger$, Han-Pang Chiu$\dagger$, Zsolt Kira*\\
*\texttt{$\{$jtian73,zkira$\}$@gatech.edu}\\ $\dagger$\texttt{$\{$niluthpol.mithun,zachary.seymour,han-pang.chiu$\}$@sri.com} 
}

\begin{document}

\maketitle
\thispagestyle{empty}
\pagestyle{empty}

\begin{abstract}
Class imbalance is a fundamental problem in computer vision applications such as semantic segmentation. Specifically, uneven class distributions in a training dataset often result in unsatisfactory performance on under-represented classes. Many works have proposed to weight the standard cross entropy loss function with pre-computed weights based on class statistics, such as the number of samples and class margins. There are two major drawbacks to these methods: 1) constantly up-weighting minority classes can introduce excessive false positives in semantic segmentation; 2) a minority class is not necessarily a \textit{hard} class. The consequence is low precision due to excessive false positives. In this regard, we propose a \textit{hard-class mining loss} by reshaping the vanilla cross entropy loss such that it weights the loss for each class dynamically based on instantaneous \textit{recall} performance. We show that the novel recall loss changes gradually between the standard cross entropy loss and the inverse frequency weighted loss. Recall loss also leads to improved mean accuracy while offering competitive mean Intersection over Union (IoU) performance. On Synthia dataset\footnote{Synthia-Sequence Summer split}, recall loss achieves $9\%$ relative improvement on mean accuracy with competitive mean IoU using DeepLab-ResNet18 compared to the cross entropy loss. Code available at \url{https://github.com/PotatoTian/recall-semseg}.
\end{abstract}


\section{Introduction}
Dataset imbalance~\cite{tian2020posterior,johnson2019survey} is an important problem for many computer vision tasks, such as semantic segmentation and image classification. In semantic segmentation, imbalance occurs as a result of natural occurrence and varying sizes of different classes. For example, in an outdoor driving segmentation dataset, light poles and pedestrians are considered minority classes compared to large classes such as building, sky, and road. These minority classes are often more important than large classes for safety reasons. Imbalance also affects the metrics used by segmentation tasks. Recent works~\cite{chen2014semantic}~\cite{chen2017rethinking}~\cite{jiang2017incorporating} use both \textit{mean} accuracy and \textit{mean}  Intersection over Union (IoU) to measure segmentation performance so that large classes do not dominate the evaluation.  While mean IoU provides a more holistic view of the performance of a model than accuracy because it considers false positive predictions (see Table~\ref{tab:metrics}), mean accuracy (also as \textit{mean} recall) does not consider the effects of false positives~\cite{grandini2020metrics}. One might argue that high recall for each class regardless of imbalance is important for safety. It can degrade segmentation quality, as shown in Fig.~\ref{fig:synthia_viz}, where small classes become indistinguishable from each other. Therefore, it is important to improve recall while maintaining a competitive mean IoU without introducing excessive false positives.

Researchers have studied the imbalance problem for classification, detection, and segmentation extensively. Most prior research has been on designing balanced loss functions. We classify existing loss functions under three categories: \textit{region-based losses}, \textit{statistics-balanced losses} and \textit{performance-balanced losses}. \textbf{Region-based losses} directly optimize region metrics (e.g., Jaccard index \cite{rahman2016optimizing}) and are mainly popular in medical segmentation applications;  \textbf{Statistics-balanced losses} (e.g., LDAM \cite{cao2019learning}, Class-Balanced (CB) loss \cite{cui2019class}) up/down weights the contribution of a class based on its class margin or class size; however, they tend to encourage excessive false positives in minority classes to improve mean accuracy especially in segmentation. A recent study in \cite{zhou2019bbn} also shows that the weighting undermines the generic representation learning capability of the feature extractors; \textbf{Performance-balanced losses} (e.g., Focal loss \cite{lin2017focal}) use a certain performance indicator to weight the loss of each class. As an example, Focal loss assumes that the difficulty of a class is correlated with imbalance and can be reflected by the predicted confidence. However, it has not been very successful in other applications (such as image classification and segmentation) for dealing with imbalance, as reported by \cite{cui2019class}. We investigate the reasons of failure in Sec.~\ref{sec:focal_analysis}.

\begin{figure}
    \centering
    \includegraphics[width=0.5\textwidth]{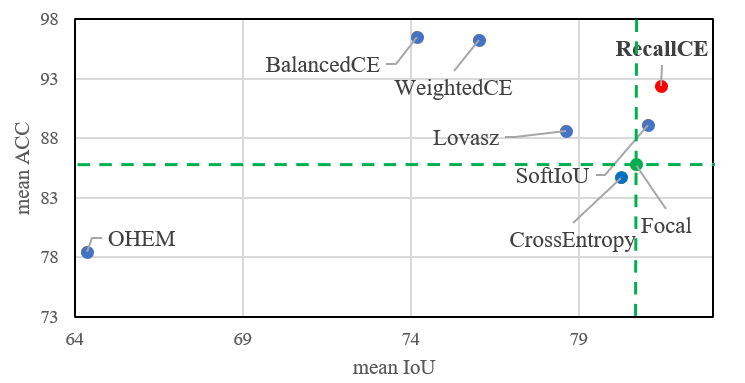}
  \caption{\textbf{Mean accuracy (ACC: $\frac{TP}{TP+FN}$) vs.  mean intersection over union (IOU: $\frac{TP}{TP+FN+FP}$) on Synthia dataset with Deeplab (Resnet18)}. Many losses improve accuracy (recall) at the expense of IoU. Only losses in the upper-right quadrant improve accuracy without introducing excessive false positives. Note: TP: true positive, FN: false negative, FP: false positive. }
  \label{fig:performance}%
\end{figure}

We propose a novel performance-balanced loss, called Recall loss, using the \textit{recall} metric to address the imbalance problem. Recall loss down/up weights a class based on the training \textit{recall} performance of that class. It is an example of \textit{hard class mining} as supposed to the hard example mining strategy in Focal loss. Unlike the statistics-balanced losses, Recall loss dynamically changes its weights with training based on per-class recall performance. The dynamism is the key to overcome many drawbacks of the statistics-balanced losses. In our experiments, the CB loss improves accuracy at the expense of Intersection over Union (IOU) which considers false positives in semantic segmentation. However, our Recall loss can effectively balance between precision and recall of each class, and hence, it improves accuracy but maintains a competitive IOU. Fig.~\ref{fig:performance} shows a mean accuracy VS. mean IoU plot for different losses on the Synthia segmentation dataset~\cite{ros2016synthia}. We use Focal loss as a reference in the figure because it does not introduce excessive false positives. While some losses provide better mean accuracy, the improvement comes at the expense of lower mean IOU. Our proposed Recall loss improves accuracy considerably while still maintaining a competitive IoU.
Further, experiments on two semantic segmentation datasets demonstrate that Recall loss shows significantly better performance than state-of-the-art loss functions in balancing recall and precision.  
\section{Related Work}

\vspace{0.1cm}
\textbf{Region-balanced Loss.}
In image segmentation, Dice and Jaccard indices (IoU) are commonly used as the evaluation metrics. However, the most common training criterion, cross entropy, does not directly optimize these metrics. In medical imaging, researchers propose to optimize soft/surrogate version of these indices. SoftIOU~\cite{rahman2016optimizing} proposes to optimize a soft version of the Jaccard index; Lovasz Softmax~\cite{berman2018lovasz} also optimizes the Jaccard index based on the Lovasz convex extension; SoftDice \cite{sudre2017generalised} and softTversky~\cite{salehi2017tversky} optimize a soft version of the Dice index and Tversky index (see Table~\ref{tab:metrics}). However, concerns have been raised in~\cite{taghanaki2019combo} on whether these losses consider the trade-off between false positives and false negatives. We show that they also tend to yield high mean accuracy at the expense of lower mean IoU.

\vspace{0.1cm}
\textbf{Statistics-balanced Loss.}
Various losses have been proposed to deal with imbalance or long-tail distributions using weighted losses. The most popular loss is the inverse-frequency loss (Weighted Cross-Entropy). It weights the cross entropy loss of each class by its inverse frequency. Class-Balanced Loss~\cite{cui2019class} motivates a weighted cross entropy loss with the concept of \textit{effective number of samples} in each class. LDAM~\cite{cao2019learning} also derives a weighted cross entropy loss based on margins between classes. Some losses require changes to the network and require an iterative optimization process, so they are not easily adaptable to semantic segmentation. We compare to the representative Class-Balanced loss in our experiments. 

\vspace{0.1cm}
\textbf{Performance-balanced Loss.}
Imbalance is also a problem in object detection, where the foreground-background imbalance is extreme and undermines learning. Online Hard Example Mining (OHEM)~\cite{shrivastava2016training} proposes to find hard examples by ranking the losses and only keeping those with the highest losses. Focal loss~\cite{lin2017focal} chooses to down weight easy samples and emphasizes hard samples by weighting each sample by $1-p$ where $p$ is the predicted probability for the sample. The weight for each sample dynamically changes with training, and the method never completely discards any samples. Focal loss is especially successful because it is easy to implement and proves effective in the binary foreground-background imbalance setting. However, it has not been very successful in other applications for dealing with imbalance, as reported by~\cite{cui2019class}.

\section{Recall Loss}

\subsection{Motivation: From Inverse Frequency Loss to Recall Loss}
\label{sec:motivation1}
To motivate our proposed loss, we first analyze the standard cross entropy loss. Let $\{x_n,y_n\} \forall n \in \{1,...,N\}$, where $x_n\in R^d, y_n \in \{1,...,C\}$ denote the set of training data and corresponding labels. Let $P_n$ denotes the predictive softmax-distribution over all classes for input $x_n$ and $P_n^i$ denotes the probability of the $i$-th class. The cross entropy loss used in multiclass classification is defined as:
\begin{align}
\label{eq:CE}
    CE &= -\sum_{n=1}^N \log (P_n^{y_n})\\\nonumber  
    &= -\sum_{c=1}^C \sum_{n:y_n=c} \log (P_n^{y_n}) = -\sum_{c=1}^C N_c \log(P^c)
\end{align}
where $\{n:y_i=c\}$ denotes the set of samples whose label is $c$ and $P^c = (\prod_{n:y_n=c}P_n^{y_n})^{1/N_c}$ denotes the \textit{geometric mean confidence} of a class $c$ and $N_c$ denotes the number of samples in that class. 
As shown in Eq.~\ref{eq:CE}, the conventional cross entropy optimizes the geometric mean confidence of each class, weighted by the number of pixels in each class. When there is a significant class imbalance in the dataset, the loss function biases towards large classes as a result of larger $N_c$.

One commonly used loss for imbalanced datasets is \textit{inverse frequency cross entropy} loss \cite{eigen2015predicting, badrinarayanan2017segnet} which assigns more weight to the loss of minority classes. Let $N$ denote the total number of pixels in the training set and $N_c$ denotes the number of pixels belonging to class $c\in\{1,..,C\}$. The frequency of a class is calculated as $freq(c) = N_c/N$. We show that while the unweighted cross entropy loss optimizes the overall confidence, the loss weighted by inverse frequency optimizes mean confidence.
If we use an inverse frequency weighting, the loss is rebalanced. Note, we leave out the $N$ in $freq(c)$ as it is shared by all classes.
\begin{align}
\label{eq:invCE}
    InvCE &= -\sum_{c=1}^C \frac{1}{freq(c)} N_c \log(P^c) \\\nonumber  
    &= -\sum_{c=1}^C \frac{1}{N_c} N_c \log(P^c)   = -\sum_{c=1}^C \log(P^c)
\end{align}

As shown in Eq.~\ref{eq:invCE}, the weighted loss optimizes the geometric mean of accuracy directly. However, the inverse frequency loss might not be optimal in practice because it over-weights the minority classes and introduces excessive false positives, i.e., it sacrifices precision for recall. This problem is especially severe in semantic segmentation \cite{chan2019application}. Applying the inverse frequency loss to segmentation increases recall for each class. However, the improvement comes at the cost of excessive false positives, especially for small classes (See Fig.~\ref{fig:synthia_viz}). 

While the inverse frequency loss solves the problem of imbalance, it focuses only on improving one aspect of the problem in classification, i.e. the recall of each class. To solve this issue, we propose to weight the inverse frequency loss in Eq.~\ref{eq:invCE} with the \textit{false negative} ($FN_c$) counts for each class. The first insight is that the $FN_c$ is bounded by the total number of samples in a class and zero, i.e.
\begin{align}
     N_c \geq FN_c \geq 0 
     \label{eq:fn}
\end{align}
By weighting the inverse frequency cross entropy loss in Eq.~\ref{eq:invCE} by the false negative counts for each class, we obtain a \textit{moderate} loss function which sits between the regular cross entropy loss and inverse frequency loss. 
\begin{align}
\label{eq:recall1}
    RecallCE &= -\sum_{c=1}^C FN_{c} \log(P^c)  = -\sum_{c=1}^C \frac{FN_{c}}{N_{c}} N_{c} \log(P^c)  \\\nonumber  
    &=-\sum_{c=1}^C \frac{FN_{c}}{FN_{c}+TP_{c}} N_{c} \log(P^c)
\end{align}
As Eq.~\ref{eq:recall1} shows, the loss can be implemented as the regular cross entropy loss weighted by class-wise \textit{false negative rate (FNR)}. The second insight is that minority classes are most likely more difficult to classify with higher FNR and large classes with smaller FNR. Therefore, similar to inverse frequency loss, gradients of minority classes will be boosted and gradients of majority classes will be suppressed. However, unlike frequency weighting, the weighting will not be as extreme as motivated in Eq.~\ref{eq:fn}.

Unlike frequency and decision margin \cite{cao2019learning} which are characteristics of the dataset, FNR is a metric of a model's performance. As we continue to update the model's parameters, FNR changes. Therefore, the weights for each class change dynamically to reflect a model's instantaneous performance. We rewrite Eq.~\ref{eq:recall1} using \textit{recall} ($1-FNR$) with a subscript $t$ to denote the time dependency. 
\begin{align}
\label{eq:recall}
    RecallCE  &= -\sum_{c=1}^C (1-\frac{TP_{c,t}}{FN_{c,t}+TP_{c,t}})  N_c \log(p^{c,t}) \\\nonumber  
    &= -\sum_{c=1}^C  \sum_{n:y_i=c}(1-\mathcal{R}_{c,t})\log(p_{n,t})
\end{align}
where $\mathcal{R}_{c,t}$ is the \textit{recall} for class $c$ at optimization step $t$.

\subsection{Gradient Analysis: Recall Loss Balances Recall and Precision}
\label{sec:gradient_analysis}
To validate the claim that recall loss balances recall and precision, we conduct a gradient analysis of the recall loss. Let's assume a binary classification task for clarity. Let $[z_1,z_2]$ and $[P_1,P_2]$ denote the pre-softmax logits and post-softmax probabilities of a classifier. We first obtain the gradient for the standard Cross-Entropy loss with respect to the logits for a single input.
\begin{align}
    \nabla_{z_i} CE &= -\log(P^y) = -\log(\frac{e^{z_y}}{\sum_i e^{z_i}}) = P_i - \mathcal{I}(y=i)
\end{align}
where $\mathcal{I}(y=i) = 1$ if $y=1$ and $0$ otherwise. Assume we have a dataset with $N_1$ and $N_2$ number of samples for each class. The gradients of the recall loss w.r.t the logit of the first class, $z_1$, is:
\begin{align}
\label{eq:recall_grad}
  & - \nabla_{z_1}\left[(1-\mathcal{R}_1)\sum_{n=1}^{N_1} \log(p_{n}^1) + (1-\mathcal{R}_2)\sum_{n=1}^{N_2} \log(p_{n}^2)\right]\\\nonumber
     & = (1-\mathcal{R}_1)\sum_{n=1}^{N_1} (p_{1,n}^{(1)} - 1) + (1-\mathcal{R}_2)\sum_{n=1}^{N_2} (p_{1,n}^{(2)}) \\\nonumber
     &= FN_1 (P_1^{(1)}-1)+ FN_2 P_1^{(2)}
\end{align}
where the superscript $(j)$ in $P^{(j)}_i$ denotes the ground truth class and the subscript denotes the class w.r.t which the gradient is calculated. $P_1^{(1)} = \frac{1}{N_1}\sum_{n=1}^{N_1}p_{1,n}^{(1)}$ denotes the average confidence of class 1 when it is the ground truth class, and $P_1^{(2)} = \frac{1}{N_2}\sum_{n=1}^{N_2}p_{1,n}^{(2)}$ denotes the average confidence of class 1 when the ground truth class is 2.

To see how the recall loss affects gradients backprobagation to the logits, we can use the fact that \textit{in a binary classification problem, a false negative (FN) in one class is a false positive (FP) in the other} to replace $FN_2$ by $FP_1$ in Eq.~\ref{eq:recall_grad}.
\begin{align}
    \nabla_{z_1} RecallCE & = FN_1 (P_1^{(1)}-1)+ FN_2 P_1^{(2)}\\\nonumber
    &=FN_1 (P_1^{(1)}-1) + FP_1 P_1^{(2)}
\end{align}
The gradient is a sum of two terms: \textbf{the first term directly encourages recall and the second term regularizes precision, i.e., reduces false positive. }
\begin{itemize}
    \item The first term $FN_1 (P_1^{(1)}-1)$ consists of gradients from samples with the ground truth class $1$. It incurs a larger \textit{negative} gradient to the class with larger false negatives ($FN$). The logit of $z_1$ will increase because we subtract the gradient in gradient descent.
    
    \item The second term $FP_1 P_1^{(2)}$ consists of the non-ground truth gradient contribution. It incurs a larger \textit{positive} gradient to the class with larger false positive ($FP$). Therefore, the logit of $z_1$ will decrease. 
\end{itemize}

\subsection{Necessity Analysis: Recall, Precision, Dice, Jaccard and Tvesky}
\label{sec:analysis_metrics}
\begin{table*}
\centering
\resizebox{1.0\linewidth}{!}{%
\begin{tabular}{c|cccccc}  
\toprule
& $Recall(\mathcal{G}_c,\mathcal{P}_c)$ & $Precision(\mathcal{G}_c,\mathcal{P}_c)$ & $Dice(\mathcal{G}_c,\mathcal{P}_c)$ & $Jaccard(\mathcal{G}_c,\mathcal{P}_c)$& $F1(\mathcal{G}_c,\mathcal{P}_c)$ & $Tversky(\mathcal{G}_c,\mathcal{P}_c)$  \\
\midrule
 Set Rep. & $\frac{|\mathcal{G}_c \cap \mathcal{P}_c|}{|\mathcal{G}_c|}$ &  $\frac{|\mathcal{G}_c \cap \mathcal{P}_c|}{|\mathcal{P}_c|}$ & $\frac{2|\mathcal{G}_i \cap \mathcal{P}_c|}{|\mathcal{P}_c|+|\mathcal{G}_c|}$ & $\frac{|\mathcal{G}_c \cap \mathcal{P}_c|}{|\mathcal{G}_c \cup \mathcal{P}_c|}$ & $\frac{|\mathcal{G}_c \cap \mathcal{P}_c|}{|\mathcal{G}_c \cup \mathcal{P}_c|+ \frac{1}{2}|\mathcal{P}_c|+ \frac{1}{2}|\mathcal{G}_c|}$ & $\frac{|\mathcal{G}_c \cap \mathcal{P}_c|}{|\mathcal{G}_c \cup \mathcal{P}_c|+ \alpha|\mathcal{P}_c|+ \beta|\mathcal{G}_c|}$ \\
 \midrule
 Boolean Rep. & $\frac{TP_c}{TP_c+FN_c}$ & $\frac{TP_c}{TP_c+FP_c}$  & $\frac{2TP_c}{2TP_c+FP_c+FN_c}$  & $\frac{TP_c}{TP_c+FP_c+FN_c}$ &$\frac{TP_c}{TP_c+ \frac{1}{2} FP_c+\frac{1}{2} FN_c}$ & $\frac{TP_c}{TP_c+\alpha FP_c+\beta FN_c}$\\
\bottomrule
\end{tabular}}
\caption{\textbf{Region Metrics:} We show both set representation and Boolean representation. TP, FN and FP stand for true positive, false negative and false positive respectively. The subscript $c$ means that the metrics are calculated for each class. }
\label{tab:metrics}
\end{table*}

Why do we not use other metrics such as F1, Dice, Jaccard and Tvesky as the weights? Following previous convention, let $\mathcal{G}_c$ and $\mathcal{P}_c$ denote the set of ground truth (positive) samples and predicted samples for the class $c$. Let $FP_c$, $TN_c$ denote the set of false positive and true negative samples respectively for class $c$, and other terms are defined similarly. 
 Recall is different from the other metrics in that it does not have false positive in the denominator and this distinction makes it ideal for weighting cross entropy loss (and others not) as shown in table~\ref{tab:metrics}. Referring back to Eq.~\ref{eq:recall}, where recall loss is defined as weighted cross entropy by $1-\mathcal{R}_c$, replacing recall by any other metrics above would result in FP appearing in the numerator of the weights. For example, a hypothetical precision loss can be defined as the following.
\begin{align}
     Precision Loss & -\sum_{c=1}^C  \sum_{n:y_i=c} \left(\frac{FP_{c,t}}{FP_{c,t}+TP_{c,t}}\right)\log(p_{n,t})
\end{align}
This formulation will introduce unexpected behavior. A large false positive count in a class will result in a large weight, which further encourages false detection for that class. This will cause the number of false positives to explode.

\section{Experiments}

\subsection{Experimental Setting}

\vspace{0.1cm}
\textbf{Datasets.} We evaluate our recall loss on two popular large-scale outdoor semantic segmentation datasets, Synthia \cite{ros2016synthia} and Cityscapes \cite{cordts2016cityscapes}. Synthia is a photorealistic synthetic dataset with different seasons, weather, and lighting conditions. Specifically, we use the Synthia-sequence Summer split for training (4400), validation (624), and testing (1272). Cityscapes consists of real photos of urban street scenes in several cities in Europe. It has 5000 annotated images for training and another 5000 for evaluation. 

\vspace{0.1cm}
\textbf{Evaluation Metrics.} We use \textit{mean accuracy} and \textit{mean IoU} as evaluation metrics, which are commonly used in prior works on semantic segmentation~\cite{chen2014semantic}~\cite{chen2017rethinking}~\cite{jiang2017incorporating}. We note that both mean accuracy and mean IoU are important metrics for semantic segmentation. While a good mean IOU indicates a balanced performance of precision and recall, mean accuracy is an indicator of the detection rate of each class, which is important for safety-critical applications such as self-driving. 

\vspace{0.1cm}
\textbf{Compared Methods.} We compare with two region-based losses, SoftIOU~\cite{rahman2016optimizing} and Lovasz softamax~\cite{berman2018lovasz}. We also compare with statistics-balanced loss including Weighted Cross-Entropy loss, also known as the Inverse-frequency (Weighted CE), Balanced Cross-Entropy loss (Balanced CE)~\cite{cui2019class}, and Online Hard Example Mining (OHEM) loss~\cite{shrivastava2016training}. We keep top $70\%$ samples in OHEM. We also compare to performance-based losses, Focal loss~\cite{lin2017focal}, in experiments. 

\vspace{0.1cm}
\textbf{Implementation.} We use DeepLabV3 \cite{chen2017rethinking} with resnet-$\{18,101\}$ \cite{he2016deep} backbones for semantic segmentation experiments. On Synthia, images are resized to 768 by 384. The resnet models are trained for 100 thousand iterations. On Cityscapes, images are resized to 769 by 769, and models are trained for 90 thousand iterations following \cite{cordts2016cityscapes}. We use the Adam optimizer \cite{kingma2014adam} with a learning rate of $10^{-3}$ and $10^{-4}$ without annealing, respectively.

\begin{figure*}
    \centering
    \includegraphics[width=1.0\textwidth]{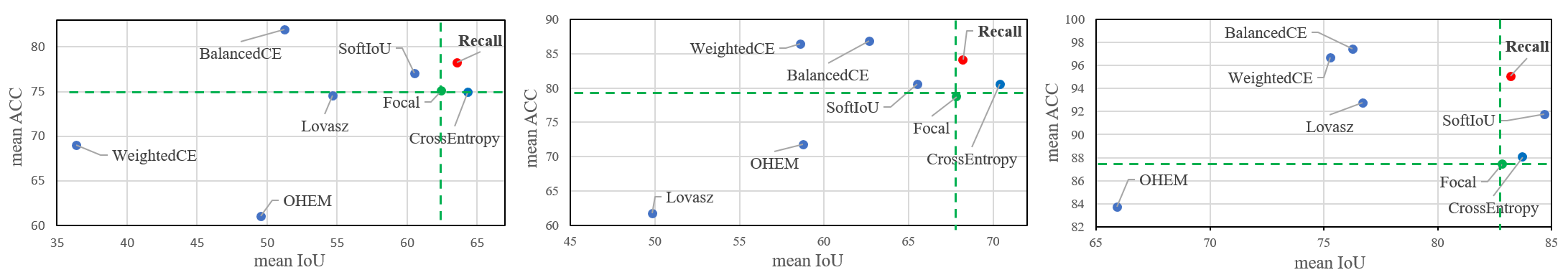}
  \caption{\textbf{Mean accuracy vs.  mean intersection over union  (all-class average)}. \textbf{Left:} Cityscapes dataset with Deeplab (Resnet18), \textbf{Middle:} Cityscapes dataset with Deeplab (Resnet101), \textbf{Right:} Synthia dataset with Deeplab (Resnet101). Only losses in the upper-right quadrant improve accuracy without introducing excessive false positives.}
  \label{fig:overall_performance}%
\end{figure*}
\subsection{Synthia}
\label{sec:synthia_exp}
\begin{figure}
    \centering
    \includegraphics[width=0.49\textwidth]{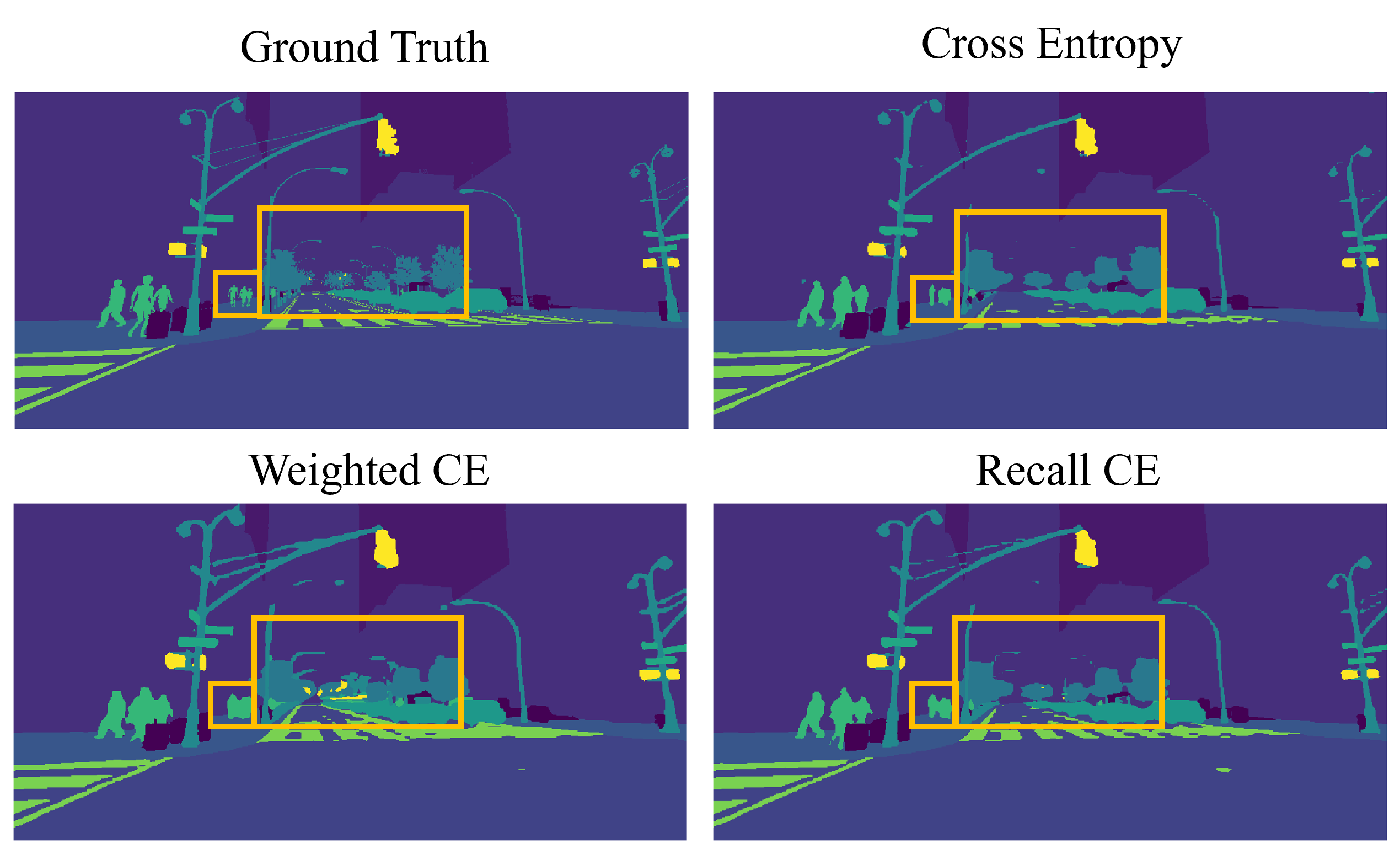}
 
  \caption{Visualization of Segmentation results on Synthia with Resnet18. Recall loss encourages models to predict more small classes such as poles, lights and pedestrians. Compared to the cross-entropy trained model, the recall loss trained model is able to output finer details especially for small classes. In contrast, Weighted CE yields excessive false positives on small classes and degrades segmentation quality significantly. (Zoom in the highlighted region for details). }
  \label{fig:synthia_viz}%
\end{figure}

We first present overall results on the synthetic Synthia segmentation dataset~\cite{ros2016synthia} using DeepLab-ResNet 101 in Fig.~\ref{fig:overall_performance}. On the plot, the x-axis represents mean IoU and the y-axis is mean accuracy. Weighted CE and Balanced CE improve the mean accuracy considerably compared to the baseline Cross-Entropy loss. However, the improvement comes at a cost of lower mean IoU. Focal loss performs similarly to the standard cross entropy Cross-Entropy loss without noticeable improvement~\cite{cao2019learning, cui2019class}.  OHEM performs worse on both metrics. We think this is because OHEM completely discards $30\%$ training samples at each iteration, and this negatively affects feature learning. SoftIoU performs reasonably well and improves both IoU and accuracy compared to the baseline. Nevertheless, Recall loss achieves the highest mean recall besides the weighted CE and Balanced CE with competitive IoU performance.  This validates our analysis that Recall loss can balance precision and recall. 

\begin{table*}[ht]
    \centering
    \resizebox{0.92\linewidth}{!}{
    \begin{tabular}{c|ccccccc|ccccccc} 
    \toprule
     &\multicolumn{7}{c}{Accuracy}& \multicolumn{7}{c}{IoU}\\
     \midrule
     & pole & sign$\dag$ & pedestrian$\dag$ & bike$\dag$ & car$\dag$ & light$\dag$ & ave.$\dag$ & pole & sign$\dag$ & pedestrian$\dag$ & bike$\dag$ & car$\dag$ & light$\dag$ & ave.$\dag$ \\
     \midrule
    CE & 98.87 &77.42 & 76.75 & 5.35 &	77.86 & 80.34 & 63.54
        & 97.62 & 70.76 & 67.53 & 5.00 & 68.98 & 70.33 & 56.5\\
    SoftIOU & 98.59 & 82.22	& 82.37 & 50.38 & 79.62 & 86.29 & 76.18
    & 97.20 & 75.70 &	70.79 &	7.07 & 69.38 & 75.94 & \underline{59.77}\\
    Balanced & 98.74 & 94.68 & 94.28 & 97.80 & 92.92 & 98.31 & \textbf{95.60}
    & 96.98 & 50.75 & 54.44 & 9.30 & 51.93 & 50.66 & 43.41\\
    
    Focal & 98.70 & 78.91 & 79.70 & 20.14 & 74.42 & 80.09 & 66.65
    &	97.48 & 71.18 & 67.53 &	14.89 &	67.41 & 69.42 & 58.09\\
    \midrule
    Recall & 98.28 & 83.94 & 87.37 & 72.69 & 89.19 & 87.24 & \underline{84.09}
    & 97.37 & 71.40 & 67.30 & 24.80 & 66.81 & 71.10 &\textbf{ 60.28}\\
    \bottomrule
    \end{tabular}}
     \caption{\textbf{DeepLab-ResNet18 per-class accuracy and IoU performance on Synthia-Summer (minority class)}. $\dag$ denotes performance on classes of IoU lower than $80$ with CE loss. Recall loss achieves the best IoU and the second best accuracy. We highlight the \textbf{Best} and \underline{Second Best} baseline method.}
    \label{tab:synthia_per_cls}
\end{table*}
In Table~\ref{tab:synthia_per_cls}, we show per-class accuracy and IoU on selected classes using DeepLab-ResNet18. Specifically, we show the results for minority classes with original IoU lower than 80 using Cross-Entropy loss. We observe that Weighted CE, Balanced CE and Recall loss can improve accuracy on small classes significantly. However, they deteriorate IoU for those classes significantly. The \textit{bike} class is the smallest class in the dataset. It only occupies $2*10^{-3}\%$ of pixels. Recall loss improves its accuracy from $5.35\%$ to $72.69\%$ while simultaneously improving IoU from $5\%$ to $24.8\%$ compared to the baseline Cross-Entropy loss. The other noticeable class is \textit{pedestrain}. Recall loss improves accuracy from $76.75\%$ to $87.37\%$ without sacrificing IoU. In contrast, Balanced loss and Weighted loss drops IoU in exchange for high accuracy.  Note that while the \textit{pole} class only occupies $1\%$ of pixels, it is an \textit{easy} class, therefore weighting is not necessary and can have a negative effect.

This observation supports our claim that Recall loss balances recall and precision because of its dynamic adaptability to performance.  We further provide a visual comparison between a model trained with the Cross-Entropy loss, Weighted CE loss and the proposed recall loss in Fig.~\ref{fig:synthia_viz}. Our method provides more fine details on small classes which are often suppressed in traditional cross entropy training. Recall loss also provides more precise segmentation than the Weighted CE loss. The Weighted CE loss enlarges all small classes and make them indistinguishable at distance.

\subsection{Cityscapes}
\renewcommand{\arraystretch}{1.2}
\begin{table*}[ht]
    \centering
    \resizebox{1.0\linewidth}{!}{
    \begin{tabular}{c|ccccccccc|ccccccccc}
    \toprule
     &\multicolumn{9}{c}{Accuracy}& \multicolumn{9}{c}{IoU}\\
     \midrule
         & person & wall$\dag$ & fence$\dag$ & pole$\dag$ & light$\dag$ & terrain$\dag$ & rider$\dag$ & mo.cycle$\dag$ & ave.$\dag$ & person & wall$\dag$ & fence$\dag$ & pole$\dag$ & light$\dag$ & terrain$\dag$ & rider$\dag$ & mo.cycle$\dag$ & ave.$\dag$\\
         \midrule
        CE & 82.98 & 69.60 & 65.63 &  58.93 & 76.76 & 62.87 & 70.16 & 67.01 & 67.63
        & 71.31 & 55.23 & 55.88 & 45.75 & 57.54 & 57.14 & 53.49 &  55.77 &  \textbf{55.23}\\
        
        SoftIOU & 85.00 & 61.23 & 68.11 &	57.70 &	79.15 &	72.53 &	71.51 &	68.57 & 68.03
        & 72.93 & 46.70 & 51.92 & 45.36 & 60.41 & 57.41 &	54.39 &	44.91 & 46.34\\
        
        Balanced & 90.07 & 61.43 & 81.13 & 71.12 &	93.36 &	83.01 &	77.90 &	84.43 & \textbf{80.88}
        & 66.48 & 45.46 & 45.82 & 40.16 &	36.58 & 48.66 & 47.62 &	38.63 & 29.58\\
        
        Focal  & 85.24 & 55.30 & 71.04 &	56.26 &	69.02 &	67.33 &	66.34 & 71.50 & 63.38
        & 70.58 & 47.91 &	55.89 &	45.33 &	56.18 &	60.02 &	50.66 & 55.17 & 51.63\\
        \midrule
        Recall  & 88.31 & 72.16 & 75.55 & 69.17 & 80.00 & 70.12 & 75.26 & 83.04 & \underline{75.06}
        & 69.06 & 57.92  & 52.50 & 44.67 & 54.64 &	57.44 & 51.14 & 41.11 & \underline{53.36} \\
        \bottomrule
    \end{tabular}}
     \caption{\textbf{DeepLab ResNet18 per-class accuracy and IoU performance on Cityscapes (minority class)}. $\dag$ denotes performance on classes of IoU lower than $80$ with CE loss. Recall loss achieves the second best IoU and accuracy. Note that balanced loss improves mean accuracy at the expense of the worst IoU. We highlight the \textbf{Best} and \underline{Second Best} method.}
    \label{tab:cityscape_per_cls}
\end{table*}

We further present results on the Cityscapes segmentation dataset~\cite{cordts2016cityscapes} with real images. Images in the dataset are real camera images and pixel-wise labels are manually assigned. Therefore, there exists more data noise as compared to the synthetic Synthia dataset. We show mean accuracy and mean IoU in Fig.~\ref{fig:overall_performance} for ResNet18 and Reset101, respectively.  Weighted CE and Balanced CE still yield the highest accuracy, but result in significantly worse IoU than baseline Cross-Entropy. Surprisingly, SoftIoU also suffers in the presence of data noise as its IoU does not match those of the Cross-Entropy loss, mostly because of lower performance on minority classes. Compared to performance on Synthia, this shows that some losses are not robust to label noise in a real dataset. Recall loss again outperforms other losses by improving mean accuracy and maintaining a good mean IOU. We further provide per-class performance on selected losses in Table~\ref{tab:cityscape_per_cls} using DeepLab-ResNet18.  Recall loss improves mean accuracy from $67.63\%$ to $75.06\%$ with minimum IoU drop on minority classes compared to Cross-Entropy. In other words, recall loss improves the detection rate of small classes such as pedestrians and light poles, while maintaining a good precision. This demonstrates the effectiveness of Recall loss on both synthetic and real outdoor segmentation datasets. No loss is perfect and improves performance on all the datasets. 

\subsection{Analysis of Focal Loss}
\label{sec:focal_analysis}
\begin{figure}
\centering
\includegraphics[width=1.0\linewidth]{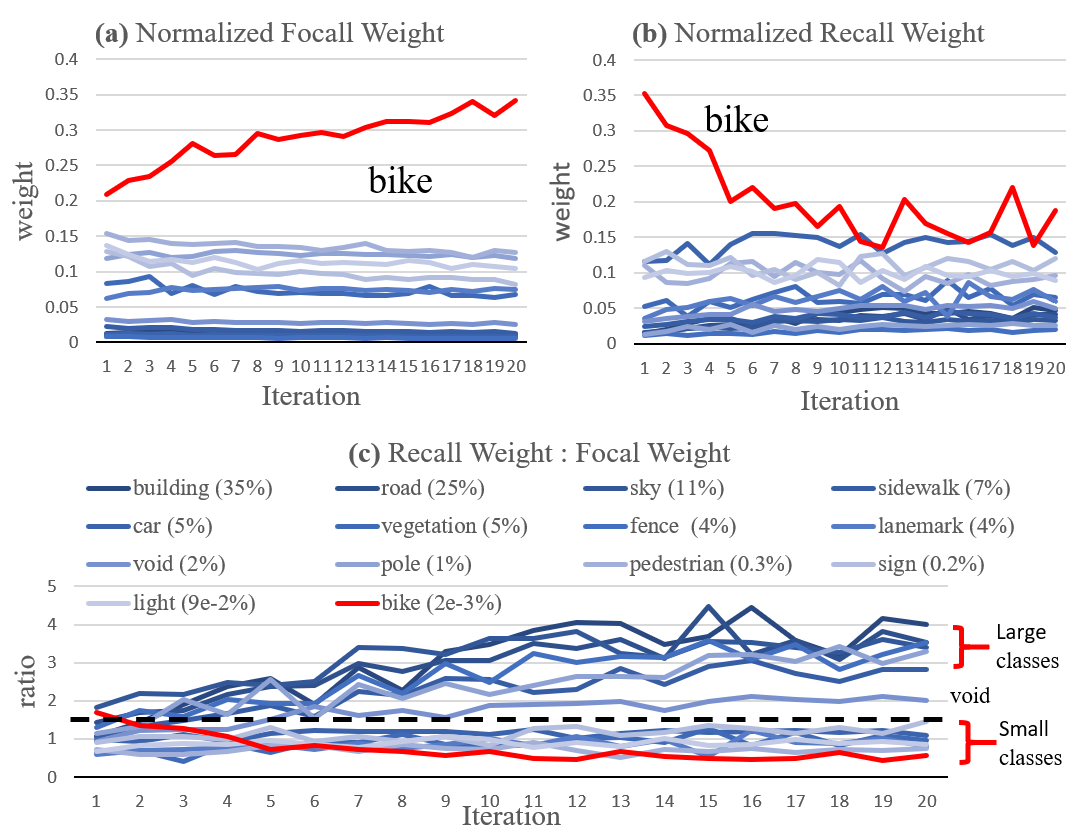}
 \caption{\textbf{(a) Normalized focal weight over iterations.} \textbf{(b) Normalized recall weight over iterations.} \textbf{(c) The ratio of recall weight and focal weight over iterations.} We present the three plots to argue that the focal loss can be minimized by making the correct predictions more confident instead of encouraging wrong predictions to become correct.} 
\label{fig:recall_focal}
\end{figure} 
Focal loss\cite{lin2017focal} is another popular and important performance-balanced loss. However, it's not specifically successful in balancing for semantic segmentation. We list Focal loss and Recall loss here for convenience.
\begin{align}
    RecallCE = -\sum_{c=1}^C  \sum_{n:y_i=c}(1-\mathcal{R}_{c,t})\log(p_{n,t}) \nonumber
\end{align}
\begin{align}
    FocalCE = -\sum_{c=1}^C\sum_{n:y_n=c}(1-p_{n,t})^\gamma\log (p_{n,t}) \nonumber
\end{align} 
where $y_n$ and $p_{n,t}$ denote the label of sample $n$ and predicted probability (confidence) for the label at time $t$. ${R}_{c,t}$ denotes the recall of class $c$ at time $t$. We use $\gamma = 1$ in following discussion. We will compare the average focal weight, $\frac{1}{N_c}\sum_{n:y_n=c}(1-p_{n,t})$, with the recall weight,$1-\mathcal{R}_{c,t}$, for each class. Because the weights are independently calculated for each class in both losses, the weights across class do not add up to one. We report \textit{normalized} weights across class instead because the relative percentage of each weight determines the focus of a loss function. The absolute weight only scales the gradient. In fig~\ref{fig:recall_focal}, we report the normalized focal weights, recall weights and recall-focal weight ratio over time during training. 

We have the following observations and follow-up discussion. 1) For Focal loss, the average bike class weight increases \textit{relatively} over time, whereas the bike class weight percentage decreases for Recall loss. This means that the focal weight $\frac{1}{N_c}\sum_{n:y_n=c}(1-p_{n,t})$ for the bike class does not decrease as fast as other classes. In other words, the confidence for the bike class does not increase \textit{relatively} to others. The recall-focal weight ratio plot tells the same story. 2) We observe that the ratios of large classes are mostly larger than one and increasing, while ratios of small classes are below one and constant. This means that Focal loss tends to assign lower and decreasing weights, which means higher and increasing confidence, to large classes.  
Colloquially, \textbf{Focal loss finds it easier to increase the confidence of large classes to reduce loss than increase the confidence of small classes}. This limits its ability to correct wrong predictions from small classes. However, because Recall loss uses the metric recall instead of predicted probability as the weights, there is no benefit to continue to increase the confidence of a sample once it is already a true positive. For Recall loss, the only way to further reduce loss is by encouraging correct predictions from small classes. The decreasing recall weight for the bike class in fig.~\ref{fig:recall_focal} shows that the performance of the bike class increases over time.  

\section{conclusion}
In this paper, we study the less-explored area of imbalance in semantic segmentation and discuss the trade-off between precision and recall, especially its asymmetric effect on large and small classes. We discovered that minority classes are more sensitive to rebalancing losses and require careful approaches to avoid excessive false positives. We propose a novel loss function based on the metric \textit{recall}. The loss function uses a \textit{hard-class mining} strategy to improve model performance on imbalanced datasets. Specifically, Recall loss weights examples in a class based on its instantaneous recall performance during  training, and the weights change dynamically to reflect \textit{relative} change in performance among classes. Experimentally, we demonstrate several advantages of the loss: 1) Recall loss improves accuracy while maintaining a competitive IoU performance. Most notably, Recall loss improves recall of minority classes without introducing excessive false positives, striking the right balance between recall and precision. 2) Recall loss avoids excessive weighting to minority but easy classes.

\section{Acknowledgement}
\label{sec:acknowledgement}
 This material is based on work supported by the Collaborative GPS-Denied Navigation for Combat Vehicles Program under Contract W9132V19C0003. We would like to thank Garry P. Glaspell, Delma B. Del Bosque, and Jean D. Nelson for their valuable feedback on the project. This work is also partially supported by ONR grant N00014-18-1-2829. Any opinions, findings, and conclusions or recommendations expressed in this material are those of the author(s) and do not necessarily reflect the views of the US government.

{\small
\bibliographystyle{IEEEtran.bst}
\bibliography{root.bib}

\begin{thebibliography}{10}
\providecommand{\url}[1]{#1}
\csname url@rmstyle\endcsname
\providecommand{\newblock}{\relax}
\providecommand{\bibinfo}[2]{#2}
\providecommand\BIBentrySTDinterwordspacing{\spaceskip=0pt\relax}
\providecommand\BIBentryALTinterwordstretchfactor{4}
\providecommand\BIBentryALTinterwordspacing{\spaceskip=\fontdimen2\font plus
\BIBentryALTinterwordstretchfactor\fontdimen3\font minus
  \fontdimen4\font\relax}
\providecommand\BIBforeignlanguage[2]{{%
\expandafter\ifx\csname l@#1\endcsname\relax
\typeout{** WARNING: IEEEtran.bst: No hyphenation pattern has been}%
\typeout{** loaded for the language `#1'. Using the pattern for}%
\typeout{** the default language instead.}%
\else
\language=\csname l@#1\endcsname
\fi
#2}}

\bibitem{tian2020posterior}
J.~Tian, Y.-C. Liu, N.~Glaser, Y.-C. Hsu, and Z.~Kira, ``Posterior
  re-calibration for imbalanced datasets,'' \emph{arXiv preprint
  arXiv:2010.11820}, 2020.

\bibitem{johnson2019survey}
J.~M. Johnson and T.~M. Khoshgoftaar, ``Survey on deep learning with class
  imbalance,'' \emph{Journal of Big Data}, vol.~6, no.~1, p.~27, 2019.

\bibitem{chen2014semantic}
L.-C. Chen, G.~Papandreou, I.~Kokkinos, K.~Murphy, and A.~L. Yuille, ``Semantic
  image segmentation with deep convolutional nets and fully connected crfs,''
  \emph{arXiv preprint arXiv:1412.7062}, 2014.

\bibitem{chen2017rethinking}
L.-C. Chen, G.~Papandreou, F.~Schroff, and H.~Adam, ``Rethinking atrous
  convolution for semantic image segmentation,'' \emph{arXiv preprint
  arXiv:1706.05587}, 2017.

\bibitem{jiang2017incorporating}
J.~Jiang, Z.~Zhang, Y.~Huang, and L.~Zheng, ``Incorporating depth into both cnn
  and crf for indoor semantic segmentation,'' in \emph{2017 8th IEEE
  International Conference on Software Engineering and Service Science
  (ICSESS)}.\hskip 1em plus 0.5em minus 0.4em\relax IEEE, 2017, pp. 525--530.

\bibitem{grandini2020metrics}
M.~Grandini, E.~Bagli, and G.~Visani, ``Metrics for multi-class classification:
  an overview,'' \emph{arXiv preprint arXiv:2008.05756}, 2020.

\bibitem{rahman2016optimizing}
M.~A. Rahman and Y.~Wang, ``Optimizing intersection-over-union in deep neural
  networks for image segmentation,'' in \emph{International symposium on visual
  computing}.\hskip 1em plus 0.5em minus 0.4em\relax Springer, 2016, pp.
  234--244.

\bibitem{cao2019learning}
K.~Cao, C.~Wei, A.~Gaidon, N.~Arechiga, and T.~Ma, ``Learning imbalanced
  datasets with label-distribution-aware margin loss,'' in \emph{Advances in
  Neural Information Processing Systems}, 2019, pp. 1565--1576.

\bibitem{cui2019class}
Y.~Cui, M.~Jia, T.-Y. Lin, Y.~Song, and S.~Belongie, ``Class-balanced loss
  based on effective number of samples,'' in \emph{Proceedings of the IEEE
  Conference on Computer Vision and Pattern Recognition}, 2019, pp. 9268--9277.

\bibitem{zhou2019bbn}
B.~Zhou, Q.~Cui, X.-S. Wei, and Z.-M. Chen, ``Bbn: Bilateral-branch network
  with cumulative learning for long-tailed visual recognition,'' \emph{CVPR},
  2020.

\bibitem{lin2017focal}
T.-Y. Lin, P.~Goyal, R.~Girshick, K.~He, and P.~Doll{\'a}r, ``Focal loss for
  dense object detection,'' in \emph{Proceedings of the IEEE international
  conference on computer vision}, 2017, pp. 2980--2988.

\bibitem{ros2016synthia}
G.~Ros, L.~Sellart, J.~Materzynska, D.~Vazquez, and A.~M. Lopez, ``The synthia
  dataset: A large collection of synthetic images for semantic segmentation of
  urban scenes,'' in \emph{Proceedings of the IEEE conference on computer
  vision and pattern recognition}, 2016, pp. 3234--3243.

\bibitem{berman2018lovasz}
M.~Berman, A.~Rannen~Triki, and M.~B. Blaschko, ``The lov{\'a}sz-softmax loss:
  A tractable surrogate for the optimization of the intersection-over-union
  measure in neural networks,'' in \emph{Proceedings of the IEEE Conference on
  Computer Vision and Pattern Recognition}, 2018, pp. 4413--4421.

\bibitem{sudre2017generalised}
C.~H. Sudre, W.~Li, T.~Vercauteren, S.~Ourselin, and M.~J. Cardoso,
  ``Generalised dice overlap as a deep learning loss function for highly
  unbalanced segmentations,'' in \emph{Deep learning in medical image analysis
  and multimodal learning for clinical decision support}.\hskip 1em plus 0.5em
  minus 0.4em\relax Springer, 2017, pp. 240--248.

\bibitem{salehi2017tversky}
S.~S.~M. Salehi, D.~Erdogmus, and A.~Gholipour, ``Tversky loss function for
  image segmentation using 3d fully convolutional deep networks,'' in
  \emph{International Workshop on Machine Learning in Medical Imaging}.\hskip
  1em plus 0.5em minus 0.4em\relax Springer, 2017, pp. 379--387.

\bibitem{taghanaki2019combo}
S.~A. Taghanaki, Y.~Zheng, S.~K. Zhou, B.~Georgescu, P.~Sharma, D.~Xu,
  D.~Comaniciu, and G.~Hamarneh, ``Combo loss: Handling input and output
  imbalance in multi-organ segmentation,'' \emph{Computerized Medical Imaging
  and Graphics}, vol.~75, pp. 24--33, 2019.

\bibitem{shrivastava2016training}
A.~Shrivastava, A.~Gupta, and R.~Girshick, ``Training region-based object
  detectors with online hard example mining,'' in \emph{Proceedings of the IEEE
  conference on computer vision and pattern recognition}, 2016, pp. 761--769.

\bibitem{eigen2015predicting}
D.~Eigen and R.~Fergus, ``Predicting depth, surface normals and semantic labels
  with a common multi-scale convolutional architecture,'' in \emph{Proceedings
  of the IEEE international conference on computer vision}, 2015, pp.
  2650--2658.

\bibitem{badrinarayanan2017segnet}
V.~Badrinarayanan, A.~Kendall, and R.~Cipolla, ``Segnet: A deep convolutional
  encoder-decoder architecture for image segmentation,'' \emph{IEEE
  transactions on pattern analysis and machine intelligence}, vol.~39, no.~12,
  pp. 2481--2495, 2017.

\bibitem{chan2019application}
R.~Chan, M.~Rottmann, F.~H{\"u}ger, P.~Schlicht, and H.~Gottschalk,
  ``Application of decision rules for handling class imbalance in semantic
  segmentation,'' \emph{arXiv preprint arXiv:1901.08394}, 2019.

\bibitem{cordts2016cityscapes}
M.~Cordts, M.~Omran, S.~Ramos, T.~Rehfeld, M.~Enzweiler, R.~Benenson,
  U.~Franke, S.~Roth, and B.~Schiele, ``The cityscapes dataset for semantic
  urban scene understanding,'' in \emph{Proceedings of the IEEE conference on
  computer vision and pattern recognition}, 2016, pp. 3213--3223.

\bibitem{he2016deep}
K.~He, X.~Zhang, S.~Ren, and J.~Sun, ``Deep residual learning for image
  recognition,'' in \emph{Proceedings of the IEEE conference on computer vision
  and pattern recognition}, 2016, pp. 770--778.

\bibitem{kingma2014adam}
D.~P. Kingma and J.~Ba, ``Adam: A method for stochastic optimization,''
  \emph{arXiv preprint arXiv:1412.6980}, 2014.

\end{thebibliography}
}

\end{document}